\documentclass[runningheads]{llncs}
\usepackage[T1]{fontenc}
\usepackage{graphicx}
\usepackage{amsmath,amssymb}

\usepackage{xcolor}
\usepackage{mathtools}
\usepackage{booktabs}
\usepackage[dvips,frame,rotate,import]{xy} 

\usepackage[noend]{algpseudocode}
\usepackage{algorithm}
\usepackage{multirow}

\DeclareMathOperator*{\argmax}{arg\; max}     % argmax

\newcommand{\Probability}{\mathbb{P}}
\newcommand{\Entropy}{H}
\newcommand{\expectedInformationGain}{I}

\newcommand{\agentIndex}{i}
\newcommand{\iterationIndex}{\agentIndex}
\newcommand{\agentOtherIndex}{j}
\newcommand{\pathIndex}{k}
\newcommand{\otherPathIndex}{j}
\newcommand{\timeIndex}{t}
\newcommand{\roundIndex}{r}

\newcommand{\baseStationCount}{m}
\newcommand{\subSearchSpaceCount}{n}
\newcommand{\mapHeight}{h}
\newcommand{\mapWidth}{w}
\newcommand{\pathLength}{\ell}

\newcommand{\searchSpace}{\mathrm{S}}

\newcommand{\genericSubSpace}{\hat{\searchSpace}}

\newcommand{\cellMapCell}{C}

\newcommand{\baseStationSet}{D}

\newcommand{\baseStation}{d}
\newcommand{\centralServer}{\sigma}

\newcommand{\agentPath}{\zeta}

\newcommand{\location}{x}

\newcommand{\environmentRandomVariable}{Z}

\newcommand{\beliefMap}[1]{\Probability(#1)}

\newcommand{\beliefRealization}{z}

\newcommand{\environmentRandomVariableSet}{\boldsymbol{Z}} 
\newcommand{\beliefRealizationSet}{\boldsymbol{z}}

\newcommand{\eventSpace}{\mathcal{Z}}

\newcommand{\hazardPresenceInCell}{\environmentRandomVariable_\cellMapCell}
\newcommand{\hazardRealizationInCell}{\beliefRealization_\cellMapCell}

\newcommand{\distruction}{\Delta}
\newcommand{\distructionRealization}{\delta}

\newcommand{\distructionSet}{\boldsymbol{\Delta}}
\newcommand{\distructionRealizationSet}{\boldsymbol{\delta}}

\newcommand{\distructionInCell}{\distruction_{\cellMapCell}}
\newcommand{\distructionRealizationInCell}{\distructionRealization_{\cellMapCell}}

\newcommand{\pathBasedSensorState}{\Theta}
\newcommand{\pathBasedSensorRealization}{\theta}

\newcommand{\distanceFunction}{f}

\newcommand{\relativeWeight}{\omega}

\newcommand{\nodeSet}{V}
\newcommand{\edgeSet}{E}
\newcommand{\timeDimension}{\mathbb{T}}

\begin{document}
\title{Bayesian Networks for Path-Based Sensors: Gathering Information and Path Planning in Communication Denied Environments}
\titlerunning{Bayesian Networks for Path-Based Sensors}
% If the paper title is too long for the running head, you can set
% an abbreviated paper title here
% \orcidID{1111-2222-(see the transcript file for additional information)</usr/share/tex3333-4444}  <-- can add this diretly after \inst{}
\author{Alkesh K. Srivastava\inst{1,2}\and
George P. Kontoudis\inst{1,3} \and
Donald Sofge\inst{1,4} \and \\
Michael Otte\inst{1}}
%
%\authorrunning{A. K. Srivastava et al.}
\authorrunning{A. K. Srivastava, G. P. Kontoudis, D. Sofge, and M. Otte}
% First names are abbreviated in the running head.
% If there are more than two authors, 'et al.' is used.
%
\institute{%
University of Maryland, College Park, MD, US.
\email{otte@umd.edu}
\and
Temple University, Philadelphia, PA, US.
\email{alkesh@temple.edu}\\
\and
Colorado School of Mines, Golden, CO, US.
\email{george.kontoudis@mines.edu}
\and
U.S. Naval Research Lab (Retired), DC, US.
\email{donald.sofge@nrl.navy.mil}
}
\maketitle              % typeset the header of the contribution
\begin{abstract}
A ``path-based sensor'' {\it produces a single observation} along a continuous path. % Alkesh's suggestion: A \textit{path-based sensor} produces a Boolean output indicating whether an event of interest occurred at any point along a continuous path, rather than at a single spatial location.
For example, a boolean path-based sensor returns a single ``1'' if an event of interest is detected at any point along the path and a ``0'' otherwise. Notably, a ``1'' provides no direct information about where along the path the event(s) may have occurred. 
Previous work has demonstrated that observations from multiple path-based sensors can be fused to create a Bayesian belief map over the spatial locations of the underlying event or phenomenon. 
Moreover, path planning can employ Shannon information theory to accelerate the rate of convergence of the belief map. 
%
% Alkesh's suggestion: 
In this paper, we present a new method to update the belief map based on a path-based sensor observation, and then plan paths to increase information gain. In contrast to prior work that approximates the posterior by averaging over the alternative event histories, we introduce a Bayesian Network (BN) formulation that models the probabilistic relationships between the latent variables and path-based sensor measurements, enabling a more principled (but still suboptimal) iterative Bayesian belief update. 
We consider static hazard detection in a communication-denied environment as a representative problem setting. The event of a robot returning from its path corresponds to a path-based hazard sensors reading of ``0'' (hazard {\it not} detected) while a robot failing to return corresponds to a reading of ``1'' (hazard detected). We consider false positives and false negatives.
%
%The new method is compared to prior work that considers both iterative single- and multi-robot deployments.
%
We find that the new method leads to quicker convergence of the belief map versus prior work in both single- and multi-robot cases.

\end{abstract}

%%%%%%%%%%%%%%%%%%%%%%%%%%%%%%%%%%%%%%%%%%%%%%%%%%%%%%%%%%%%%%%%%%%%%%%%%%%%%%%%
%---------------------------------------
%---------------------------------------
%---------------------------------------
\section{Introduction}
\label{sec:introduction}
%---------------------------------------
%---------------------------------------
%---------------------------------------

\begin{figure*}[!t]
\vspace{.12cm}

\begin{xy}
  \xyimport(100,100){\includegraphics[width=\textwidth,height=5.7cm]{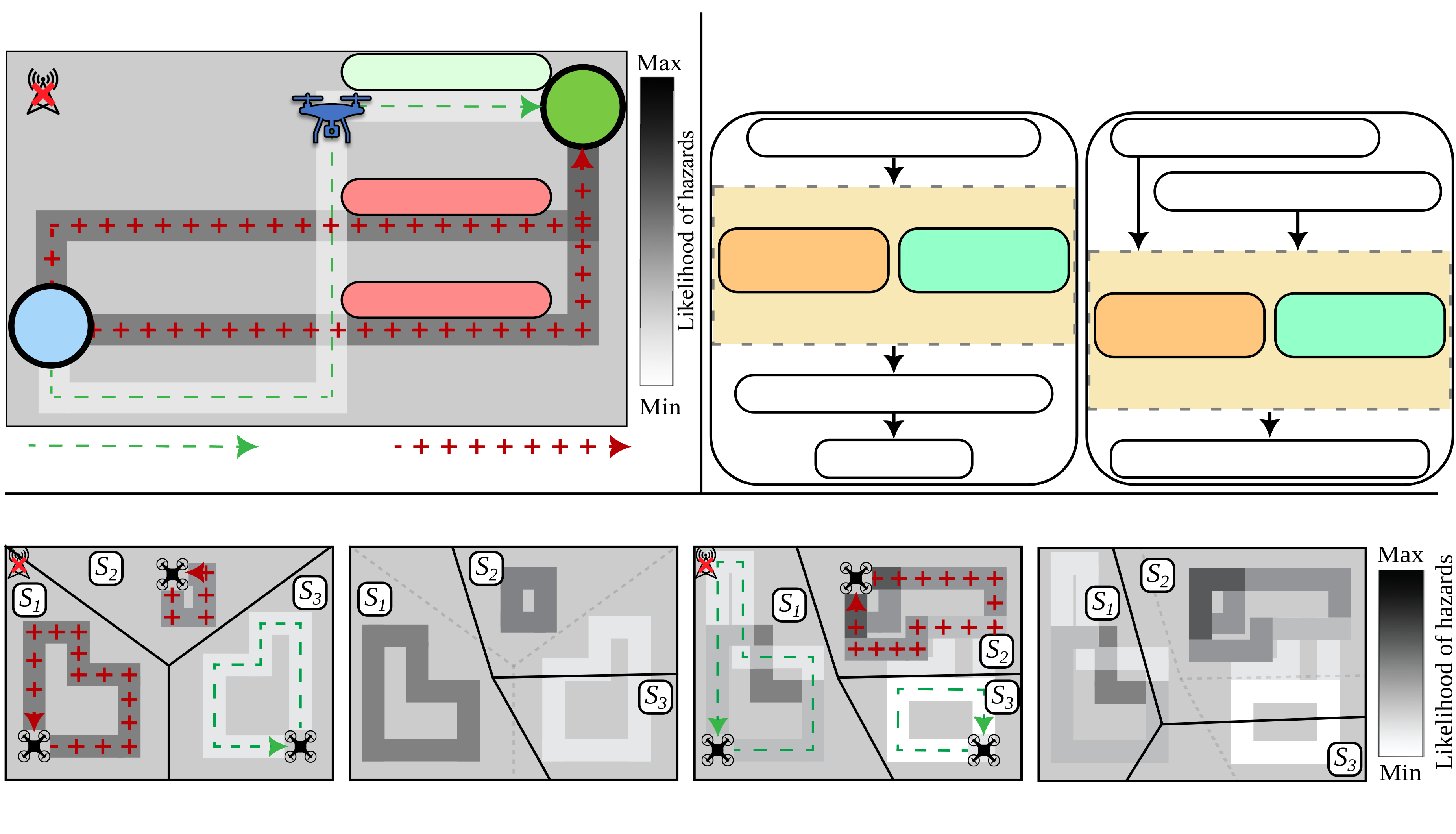}}
     ,(24,97.75)*{\text{Path-Based Sensor Concept from \cite{otte2021path}}}
     ,(76,97.75)*{\text{Key Differences vs.\ \cite{otte2021path,Srivastava.etal.DARS22}}}
     ,(10.00,85.50)*{\text{\tiny No Communication}}
     ,(30.65,91.10)*{\text{\scriptsize Path 3, $\!{\pathBasedSensorState\!=\!0}$}}
     ,(30.65,76.10)*{\text{\scriptsize Path 2, $\!{\pathBasedSensorState\!=\!1}$}}
     ,(30.65,63.70)*{\text{\scriptsize Path 1, $\!{\pathBasedSensorState\!=\!1}$}}
     ,(3.5,61.00)*{\text{\tiny Start}} 
     ,(40,87.30)*{\text{\tiny Goal}} 
     ,(9.5,42.9)*{\text{\scriptsize Agent Survived}}    
     ,(35,42.9)*{\text{\scriptsize Agent Destroyed}}    
     ,(61.5,93)*{\text{\scriptsize{Calculation of path $\agentPath$}}}   
     ,(61.5,89)*{\text{\scriptsize{given prior belief map}}}   
     ,(61.5,83.40)*{\text{\scriptsize{Prior Belief Map}}}   
     ,(61.5,75.25)*{\text{\scriptsize\textit{\textbf{Inference mechanism}}}}
     ,(55.00,70.25)*{\text{\scriptsize{Weighted}}} 
     ,(55.00,67.25)*{\text{\scriptsize{Average}}}      
     ,(55.00,63.25)*{\text{\tiny\textbf{\textit{Used in}}}}
     ,(55.00,60.50)*{\text{\tiny\textbf{\textit{\cite{otte2021path,Srivastava.etal.DARS22}}}}}
     ,(67.75,70.25)*{\text{\scriptsize{Bayesian}}} 
     ,(67.75,67.25)*{\text{\scriptsize{Network}}}  
     ,(67.75,63.25)*{\text{\tiny\textbf{\textit{Proposed}}}}
     ,(67.75,60.50)*{\text{\tiny\textbf{\textit{Method}}}}
     ,(61.5,52.50)*{\text{\scriptsize{Expected Info. Gain}}}   
     ,(61.5,44.95)*{\text{\scriptsize{Path $\agentPath$}}}   
     ,(87,93)*{\text{\scriptsize{Calculation of posterior}}}   
     ,(87,89)*{\text{\scriptsize{belief map given $\pathBasedSensorState = \pathBasedSensorRealization$}}}   
%     ,(87,89.0)*{\text{\scriptsize{observation along path}}}   
%
     ,(85.5,83.40)*{\text{\scriptsize{Observation $\pathBasedSensorRealization$}}}   
     ,(88.75,76.75)*{\text{\scriptsize{Prior Belief Map}}}  
     ,(87,67.25)*{\text{\scriptsize\textit{\textbf{Inference mechanism}}}}
     ,(81.00,62.25)*{\text{\scriptsize{Weighted}}} 
     ,(81.00,59.25)*{\text{\scriptsize{Average}}}      
     ,(81.00,55.25)*{\text{\tiny\textbf{\textit{Used in}}}}
     ,(81.00,52.50)*{\text{\tiny\textbf{\textit{\cite{otte2021path,Srivastava.etal.DARS22}}}}}
     ,(93.75,62.25)*{\text{\scriptsize{Bayesian}}} 
     ,(93.75,59.25)*{\text{\scriptsize{Network}}}  
     ,(93.75,55.25)*{\text{\tiny\textbf{\textit{Proposed}}}}
     ,(93.75,52.50)*{\text{\tiny\textbf{\textit{Method}}}}
     ,(87,44.95)*{\text{\scriptsize{Posterior Belief Map}}}  
     ,(50,37.5)*{\text{Multiple Simultaneous Path-Based Sensors in Entropy-Weighted Voronoi Regions \cite{Srivastava.etal.DARS22}}}  
\end{xy}

\vspace{-.55cm}

\noindent\begin{minipage}[t]{2.7cm}
\scriptsize 
(a) Agents plan and traverse paths in their search spaces.
\end{minipage}
\hspace{.01cm}
\begin{minipage}[t]{2.7cm}
\scriptsize 
(b) New computation of entropy-weighted Voronoi partitions
\end{minipage}
\hspace{.01cm}
\begin{minipage}[t]{2.7cm}
\scriptsize 
(c) Agents plan and traverse paths in their search spaces.
\end{minipage}
\hspace{.01cm}
\begin{minipage}[t]{2.7cm}
\scriptsize 
(d) New computation of entropy-weighted Voronoi partitions
\end{minipage}
    
    \caption{{\bf Top-Left:} Belief about hazard existence (grayscale) from three path-based sensor observations. Agents are destroyed along paths 1 and 2 (red), causing two sensor triggers (${\pathBasedSensorState=1}$) that increase belief about hazards along those paths (dark gray). An agent survives path 3 (light green) causing a non-trigger (${\pathBasedSensorState=0}$) decreasing belief about hazards (light gray). {\bf Bottom:} Extension of this idea to multiple simultaneous agent deployment, each in its own entropy-weighted Voronoi partition. {\bf Top-Right:} Conceptual framework highlighting the key difference  between the approach presented in this paper using a Bayesian Network (green) versus prior work~\cite{otte2021path,Srivastava.etal.DARS22}~(orange).}
    \label{fig:concept_image}  
\end{figure*}

Unlike conventional sensors, which take measurements or observations from a single point or a sensor footprint, a {\it path-based sensor} takes a single measurement or observation along a trajectory. A boolean path-based sensor returns a ``1'' or ``0,'' respectively, depending on whether or not an event of interest is observed (at least once) along an entire trajectory.
As an illustrative example, consider an autonomous underwater robot that slowly collects water in a tank as it travels between a start and goal location. After the mission, the collected water is analyzed in a laboratory to determine whether a particular microorganism is present. While it is possible to determine (after the analysis) that one or more microorganisms have been collected, the exact location(s) of collection are not directly observable.
Assuming that the habitats of the microorganisms are known to be localized to a small number of (initially unknown) regions within the environment, we can use multiple robot deployments and laboratory tests to identify where the habitat regions are located. In particular, we can refine a Bayesian belief map of microorganism habitat across the environment while deploying our robot (or even multiple robots) in an iterative fashion.
Moreover,  if the robot’s path can be designed, then we can optimize each path, based on the current belief map, to maximize information gain about habitat existence.

The general concept of path-based sensors is introduced in~\cite{otte2021path}, where it is proposed as a solution to the problem of gathering information about hazards in a communication-denied environment. Expendable agents are sent along predefined paths through the environment and their survival versus apparent destruction is used to gather information about hazards. A path-based sensor reading of ``0'' (hazard not detected) occurs each time an agent successfully completes its path. A path-based sensor reading of ``1'' (hazard detected) occurs each time an agent does {\it not} return, i.e., such that it is assumed to have been destroyed somewhere {\it en route}.
In \cite{otte2021path}, an approximate Bayesian belief map update is presented, as well as a path-planning algorithm that leverages this update within a Shannon information-based framework. The related problem of target search in a hazardous communication-denied environment is also considered --- where targets, e.g., search and rescue victims, are different from hazards.

Other work on path-based sensors has considered deploying multiple homogeneous agents~\cite{Srivastava.etal.DARS22} and heterogeneous agents~\cite{srivastava2025behaviorally} simultaneously, having a small finite number of agents \cite{McGuire.etal.IROS24} or path attempts \cite{MAYHUGH.masters}, and cases in which deviation from the intended path is beneficial \cite{Mendelsohn.etal.AAMAS24}. 
Any iterative path-based sensor update that retains (only) a fixed amount of data (like a map) is necessarily suboptimal after a certain number of iterations \cite{MAYHUGH.masters} (and the current paper is no exception); achieving optimality requires saving all path-based sensor readings, and then considering all interactions among them when calculating each next belief map\footnote{The authors credit Dylan Shell for sharing this insight with us, and thank him for thoughtful discussion on the topic during the WAFR'26 Birds of a Feather Session.}. This quickly becomes intractable. 
In the event that a path-based sensor is triggered,
prior work relies on an approximation 
%to calculate the posterior belief map 
in which the posterior belief maps of mutually exclusive histories (of where the agent may have been destroyed) are calculated separately, and then combined based on relative likelihood as estimated by the prior belief map. 
Our preliminary investigation (see technical report \cite{srivastava.ArXive23}) indicated that a better approximation is possible by considering the correlations that exist between the alternative histories.

\textbf{\textit{Contribution:}}~% of the current paper:}} 
We replace the approximation used in \cite{otte2021path,Srivastava.etal.DARS22,McGuire.etal.IROS24,Mendelsohn.etal.AAMAS24,srivastava2025behaviorally}
with a Bayesian Network (BN) representation of  a path-based sensor that considers a comprehensive joint probability space, iteratively accounting for static hazards, absence of hazards, agent survival, and non-survival along each path.
%
% Otte moved this to the related work section--> The current paper extends our preliminary work, i.e., that appeared in a non-archival workshop \cite{Srivastava.etal.ICRA_CCMRSPCL}  and technical report \cite{srivastava.ArXive23}), by providing a more rigorous algorithmic formulation, expanded experiments, deeper discussion of the results.
%
We show how the new update can be used within the iterative $n$-robot-at-a-time problem variant. 
%
%Note that we consider the case of hazard detection in a communication denied environment, and prohibit agent-to-base as well as inter-agent communication. 
%
We empirically compare the new method to prior work \cite{otte2021path,Srivastava.etal.DARS22} using simulations.
We find that the new method improves the accuracy of the resulting Bayesian updates, %calculations, 
accelerates convergence of the Bayesian belief map, and increases resilience to noisy data (false positives and false negatives) compared to~\cite{otte2021path} for single-robot and %compared to~
\cite{Srivastava.etal.DARS22} for parallel deployment of multiple robots.

This paper is organized as follows.
Related work is discussed in Section~\ref{sec:rw}.
The path-based sensor problem is formulated in Section~\ref{sec:problem_formulation} and 
the approaches to solving the problem are discussed in Section~\ref{sec:approach}. 
%; the old approach used in prior work is outlined in Section~\ref{subsec:approach_old}, and the Bayesian Network approach is discussed in Section~\ref{subsec:pbs-bn} (single robot) \ref{subsec:approach_multi} (multi-robot). 
Algorithms, Simulation experiments (and their results), and conclusions appear in Sections~\ref{sec:algorithms}-\ref{sec:conclusion}, respectively.

%---------------------------------------
%---------------------------------------
\section{Related Work}
\label{sec:rw}

Bayesian frameworks have been extensively used in robotics for target estimation~\cite{tomo}, where the posterior estimates of targets after observation of an event are more informative than the prior. Information theory~\cite{shannon1948mathematical} serves as the foundation for information-theoretic planners. The connection between Bayesian inference and information theory~%, presented in 
\cite{oladyshkin2019connection} has been foundational in the development of information-theoretic planners and inverse modeling problems~\cite{mohammad2015entropy}. Mutual information, a measure of relative information, has been used %as a measure for controlling agents 
to improve estimation~\cite{charrow2014approximate}, for target tracking~\cite{grocholsky2002information}, and exploration~\cite{bourgault2002information}.

%Estimating the location of environmental hazards in lethally hostile environments can help to ensure human safety and is a prerequisite for avoiding and removing hazards. 
%is a crucial problem within the robotics community. 

With the increasing availability and reliability of autonomous systems, there has been a growing emphasis on utilizing multi-robot %agent robotic 
systems to tackle the problem of information gathering. 
Accurate and rapid estimation of hazard location was investigated initially in~\cite{schwager2017multi}. However, unlike the problem we consider, inter-agent communications %with the agent 
are allowed in \cite{schwager2017multi}, which facilitates instant identification of false positives. While some prior work has considered multi-robot systems in environments where communication is strictly prohibited~\cite{gielis2022critical,Srivastava.etal.DARS22}, other work~\cite{flint2003cooperative,sato2010path,yang2007multi,lyu2016k,jorgensen2017matroid,yang2004decentralized} considers localizing hazards in environments with limited communication. For example, %Julian~\emph{et al.}~
\cite{julian2012distributed} provides a formal derivation of the gradient of mutual information, demonstrating that under a multi-agent gradient ascent control strategy, information entropy would %eventually 
approach zero at the limit. %as the time approaches infinity. 
The latter formulations %are able to assume 
consider that the observations of interest are explicitly provided in real-time, while we account for an implicit type of path-based~sensor observations that occur after each path has been traversed.

Probabilistic graphical models %are described as a way to 
are used to represent the conditional dependencies among the involved random variables~\cite{pearl1988probabilistic}. Bayesian networks have been used in numerous applications such as bio-informatics~\cite{yu2004advances}, machine learning classifiers~\cite{friedman1997bayesian}, computer vision~\cite{nie2018deep}, and deep learning~\cite{hao2016towards}. In the domain of environment estimation, a probabilistic graphical model-based method is used for tracking targets in~\cite{uney2007graphical} and \cite{uney2008target}. We seek to %The work presented in this paper introduces techniques to 
detect hazards by estimating the origin of path-based sensor realizations using probabilistic graphical modeling. %This approach has not yet been studied in the context of path-based sensors.

% In our previous work~\cite{otte2021path}, we introduce the concept of path-based sensor observations as a solution to the %the state-of-the-art solution for comprehensive joint p
% information gathering problem in communication-denied environments. 
Path-based sensors are theoretical sensors that provide binary observations to indicate the occurrence of an event along a traversed path, without reporting precise event locations~\cite{otte2021path}. These sensor observations are particularly relevant in scenarios where the occurrence of an event can only be determined in post-processing. For example, when wet-lab analysis is required to determine whether a chemical or biological agent was encountered, and when robots are deployed to detect static hazards in a communication-denied environment \cite{otte2021path,Srivastava.etal.DARS22,McGuire.etal.IROS24,Mendelsohn.etal.AAMAS24}.

The current paper goes beyond prior work on path-based sensors by exploring a reformulation of the iterative map update procedure using tools from Bayesian networks. 
Some of the ideas in the current paper were presented in a non-archival workshop \cite{Srivastava.etal.ICRA_CCMRSPCL} and also documented in a technical report \cite{srivastava.ArXive23}. The current paper goes beyond  \cite{Srivastava.etal.ICRA_CCMRSPCL,srivastava.ArXive23} by providing a more rigorous algorithmic formulation, expanded experiments, deeper discussion of the results, as well as showing how the new update can be used within the iterative $n$-robot-at-a-time problem variant.

%---------------------------------------
%---------------------------------------
%---------------------------------------
\section{Problem Formulation}
\label{sec:problem_formulation}
%---------------------------------------
%---------------------------------------
%---------------------------------------
We consider a search space~$\searchSpace\subset\mathbb{R}^2$ composed of a $\mapHeight \times \mapWidth$ grid of discrete cells. Let $\cellMapCell$ denote a single cell within the grid. 
Hazardous elements are present in $\searchSpace$. 
We use $\environmentRandomVariable$ to denote the state of the environment with respect to hazards. $\environmentRandomVariable$ is a discrete time random variable that takes values on the alphabet $\eventSpace = \{0,1\}^{\mapHeight \times \mapWidth}$; in other words, $\eventSpace$ is an event space that contains each of the $2^{\mapHeight \times \mapWidth}$ different possibilities of having a hazard $\{1\}$ or not $\{0\}$ in each of the $\mapHeight \times \mapWidth$ separate cells of $\searchSpace$. For tractability, we assume that hazard states are static and statistically independent across cells; modeling spatial or temporal dependencies would introduce additional edges in the Bayesian network and significantly increase the complexity of inference.
Our belief $\beliefMap{\environmentRandomVariable}$ of hazard state is updated after each search round. When necessary for the sake of clarity, we use superscripts, e.g., $\environmentRandomVariable^{(\iterationIndex)}$, to indicate the belief map that incorporates information from search rounds $1$ through $\iterationIndex$. Our prior belief of hazard state prior to any search round is denoted  $\environmentRandomVariable^{(0)}$.
Hazard state in cell $\cellMapCell$ is denoted $\hazardPresenceInCell$ and hazard state in the subspace $\genericSubSpace \subset \searchSpace$ is denoted $\environmentRandomVariable_{\genericSubSpace}$.
Similarly, our beliefs regarding these discrete time random variables are denoted 
$\beliefMap{\hazardPresenceInCell = \hazardRealizationInCell}$ and $\beliefMap{\environmentRandomVariable_{\genericSubSpace} = \beliefRealization_{\genericSubSpace}}$, respectively.

We assume that hazards can only be detected indirectly when agents fail to return after a search round, and that the destruction of agents cannot be observed directly. 
Even though hazard presence cannot be observed directly, it is convenient to let $\hazardPresenceInCell = 1$ and $\hazardPresenceInCell = 0$ denote the existence or absence of a hazard in cell $\cellMapCell$, respectively.
We use $\distructionInCell$ to denote the random variable of destruction in a given cell~$\cellMapCell$, i.e. $\distructionInCell = 1$ denotes that the agent was destroyed in cell $\cellMapCell$ whereas $\distructionInCell = 0$ denotes that the agent survived a passage through cell $\cellMapCell$. We use the variable $\distructionRealizationInCell \in \{0, 1\}$ to denote the realization of such an event.

Agents are deployed to gather information about hazards 
by attempting to traverse paths through the environment. 
The environment contains a set of $\baseStationCount$~base~stations~$\baseStationSet = \{\baseStation_1, \baseStation_2, \hdots, \baseStation_{\baseStationCount} \}\subset\searchSpace$.
Agents start and end their search rounds at their designated base stations and 
cannot communicate from the field {\it during} a path traversal. However, reliable communication is assumed to exist among base stations %all base stations have reliable communication with other base stations 
and with a central server~$\centralServer$. Thus, knowledge about whether an 
agent $\agentIndex$ survives its path $\agentPath_{\agentIndex}$ is stored at $\centralServer$ and available at all base stations.
%
%The time it takes for agent $\agentIndex$ to transition from cell $\cellMapCell_{\pathIndex}$ to adjacent cell $\cellMapCell_{\pathIndex+1}$ is defined as timestep $\timeIndex$. \todo{CHECK IF $\timeIndex$ IS USED.}

\begin{definition}\label{def:path_based_sensor}
\textit{Path-based sensor}: A binary sensor that reports the occurrence of an event along a path but does not provide the location (i.e., cell) of occurrence. 
\end{definition}

The random variable associated with an agent $\agentIndex$ surviving a path $\agentPath_{\agentIndex}$ is denoted $\pathBasedSensorState_{\agentIndex}$.
The destruction or survival of agent $\agentIndex$ along path $\agentPath_{\agentIndex}$ is a path-based sensor (Definition~\ref{def:path_based_sensor}) observation ${\pathBasedSensorRealization_{\agentIndex} \in \{1, 0\}}$, where ${\pathBasedSensorRealization_{\agentIndex} = 0}$ is survival (hazard not observed) and ${\pathBasedSensorRealization_{\agentIndex} = 1}$ is destruction (hazard observed).
Belief of path survival is thus $\Probability(\pathBasedSensorState_{\agentIndex} = \pathBasedSensorRealization_{\agentIndex})$. 
We consider cases where the path-based sensor may report false positives 
or false negatives. % triggering. 
A false-positive accounts for a faulty agent that
fails to traverse a path regardless of the presence of hazards, whereas a false-negative accounts for the chance of an agent surviving a hazardous cell.

The Shannon information entropy of the belief of hazard existence in cell ${\cellMapCell}$ is defined as~$\Entropy(\hazardPresenceInCell) \coloneqq -\sum_{\hazardRealizationInCell \in \{0,1\}}\Probability(\hazardPresenceInCell = \hazardRealizationInCell)\log\Probability(\hazardPresenceInCell = \hazardRealizationInCell)$.
Similarly, $\Entropy(\environmentRandomVariable_{\genericSubSpace})$ is the Shannon information entropy of the entire subspace $\genericSubSpace$.
As in \cite{otte2021path}, we assume that hazard effects are local to each cell. This enables us to calculate $\Entropy(\environmentRandomVariable_{\genericSubSpace})$ in $\mathcal{O}(\mapHeight \times \mapWidth)$ time by summing over the entropy of cells $\cellMapCell \in \genericSubSpace$,  instead of the $\mathcal{O}(2^{\mapHeight \times \mapWidth})$ time required, in general, to consider all ${\environmentRandomVariable \in \eventSpace}$.
\begin{equation*}
\Entropy(\environmentRandomVariable_{\genericSubSpace}) 
= \sum_{\cellMapCell \subset \genericSubSpace}\Entropy(\environmentRandomVariable_{\cellMapCell})
= -\sum_{\cellMapCell \subset \genericSubSpace}
\sum_{\hazardRealizationInCell \in \{0,1\}}\Probability(\hazardPresenceInCell = \hazardRealizationInCell)\log\Probability(\hazardPresenceInCell = \hazardRealizationInCell).
\end{equation*}
Observe that a path $\agentPath$ traverses cells in the order $\cellMapCell_1, \ldots, \cellMapCell_\pathLength$ and revisiting cells is allowed such that $\cellMapCell_\pathIndex$ and $\cellMapCell_\otherPathIndex$ may or may not be the same. For ease of presentation, we abuse the notation and allow a path to be interpreted as the sequence of cells through which it traverses $\agentPath = \langle \cellMapCell_1, \ldots, \cellMapCell_\pathLength \rangle$. 
The expected information gain about the presence of hazards can only be obtained indirectly through path-based sensor observations indicating agent destruction. Given a sensor observation $\pathBasedSensorState_\agentIndex$, the information gain is defined as the expected reduction in the entropy associated with $\environmentRandomVariable$,
$$\expectedInformationGain(\environmentRandomVariable^{(\agentIndex)};\pathBasedSensorState_\agentIndex) = 
\Entropy(\environmentRandomVariable^{(\agentIndex - 1)}) - \Entropy(\environmentRandomVariable^{(\agentIndex)}\, | \,\pathBasedSensorState_\agentIndex, \environmentRandomVariable^{(\agentIndex - 1)}) ,
$$
where 
$\Entropy(\environmentRandomVariable^{(\agentIndex)}\, | \,\pathBasedSensorState_\agentIndex, \environmentRandomVariable^{(\agentIndex - 1)})
= \sum_{\pathBasedSensorRealization_\agentIndex \in \{0, 1\}} \Probability(\pathBasedSensorState_\agentIndex = \pathBasedSensorRealization_\agentIndex)\Entropy(\environmentRandomVariable^{(\agentIndex)}\, | \, \pathBasedSensorState_\agentIndex = \pathBasedSensorRealization_\agentIndex,  \environmentRandomVariable^{(\agentIndex - 1)})
$ 
is the conditional entropy computed by summing over the possible outcomes of ${\pathBasedSensorState_\agentIndex = \pathBasedSensorRealization_\agentIndex}$ weighted by our current belief of their probability of occurring.

\subsection{One Agent per Iteration Problem Formulation}

In the classical version of the problem \cite{otte2021path}, agents are deployed iteratively, one-at-a-time, to gather information from the search space~$\searchSpace$. In the one-at-a-time scenario, we use the index $\agentIndex$ to indicate the agent that is deployed at iteration $\agentIndex$.
Agent $\agentIndex $ has knowledge of the survival status of all agents $1, \ldots, \agentIndex - 1$ and up-to-date access to the corresponding hazard belief  $\beliefMap{\environmentRandomVariable^{(\agentIndex - 1)}}$ that reflects this knowledge. We now formalize the iterative version of the problem.

\begin{problem}
\label{ps:problem} 
\textit{Iterative path-based sensing for hazard detection in communication-denied environments}:
Given a search space $\searchSpace$ in which communication is prohibited from the field and where $\searchSpace$ contains an unknown number of hazards and $\baseStationCount$ base stations, hazards can only be detected indirectly through the act of destroying agents, and assuming that the base stations can communicate with a central server $\centralServer$, then iteratively deploy agents along paths $\agentPath_1, \ldots, \agentPath_{\agentIndex - 1}$ and observe whether or not the agents survive to gather information about hazards to refine hazard belief $\beliefMap{\environmentRandomVariable}$. 
\end{problem}

Assuming that we can choose the paths that agents traverse, the problem of path optimization can also be formalized.

\begin{problem}
\label{ps:optimization} 
\textit{Iterative information-optimal path-based sensor shape design}:
Given a prior belief $\beliefMap{\environmentRandomVariable}$ calculated from path-based sensor readings associated with $\agentPath_1, \ldots, \agentPath_{\agentIndex - 1}$,  calculate a new path~$\agentPath^{\star}$ for agent $\agentIndex$ of length $\pathLength_{\agentIndex}$ that maximizes the expected information  gained from the path-based sensor $\pathBasedSensorState$ that will be generated by agent $\agentIndex$ following path $\agentPath_{\agentIndex}$, %i.e., 
\begin{equation}
     \agentPath^{\star} = \argmax_{\agentPath_\agentIndex} \expectedInformationGain(\environmentRandomVariable; \pathBasedSensorState_\agentIndex).
\end{equation}
\end{problem}

\subsection{$\subSearchSpaceCount$-Agents per Iteration Problem Formulation}

In \cite{Srivastava.etal.DARS22},~$\subSearchSpaceCount$ separate agents are deployed simultaneously during each iteration; however, agents are deployed in non-overlapping regions for the sake of parallel actuation as well as the convenience of the individual path calculations. In this multi-agent per iteration case, there is no longer a one-to-one mapping between iteration index and agent index. So, in the multi-agent case, we will refer to such iterations as {\it rounds} and use $\roundIndex$ to denote the current round. While $\agentIndex$ is still used to track agents and their paths, it now refers to the $\agentIndex$-th agent that is deployed in a particular search round $\roundIndex$.  
During each round $\roundIndex$, the map is partitioned into $\subSearchSpaceCount$ regions of approximately equal entropy corresponding to each agent. Each region contains at least one base station. The map partitioning changes as $\roundIndex$ increases due to changing information value on %the fact that more or less information will be gathered from 
different cells.

During round $\roundIndex$,  each agent $\agentIndex$ starts at base station $\baseStation_{\agentIndex}^{(\roundIndex)}$ and is assigned the corresponding partition (subregion) as its search space $\searchSpace_{\agentIndex}^{(\roundIndex)}  \subset \searchSpace$. In each round $\roundIndex$, the union of all partitions forms the entire search space $\bigcup_{\agentIndex=1}^{\subSearchSpaceCount}\searchSpace_{\agentIndex}^{(\roundIndex)}  = \searchSpace$, and there is no overlap between partitions, $\searchSpace_{\agentIndex}^{(\roundIndex)}  \cap \searchSpace_{\agentOtherIndex}^{(\roundIndex)}  = \emptyset$ for ${\agentIndex \neq \agentOtherIndex}$.
An agent is not expected to search the entire space in a single round, rather, it considers only cells within its search space partition when calculating its path $\agentPath_{\agentIndex}^{(\roundIndex)}  \subseteq \searchSpace_{\agentIndex}^{(\roundIndex)} $.

Each agent $\agentIndex$ is capable of visiting %up to 
$\pathLength_{\agentIndex}$ cells within its assigned search space, forming a path $\agentPath_{\agentIndex}^{(\roundIndex)}  = \langle \cellMapCell_{1},\cellMapCell_{2},\ldots,\cellMapCell_{\pathLength_{\agentIndex}}\rangle\subseteq~\searchSpace_{\agentIndex}^{(\roundIndex)}$. Note that in our experiments, we consider the case where all agents have an identical path length constraint $\pathLength_\agentIndex =  \pathLength$.
We assume that there is no communication between the centralized server~$\centralServer$ and the agents about their whereabouts until the search round~$\roundIndex$ concludes.

%At the end of each exploration, when all surviving agents return to their respective base stations, a search round $r$ is completed. 

\begin{problem}
\label{ps:multi} 
\textit{Simultaneous iterative path-based sensing for hazard detection in communication-denied environments with simultaneous deployments}:
Given a search space $\searchSpace$ in which communication is prohibited from the field that contains an unknown number of hazards and where $\searchSpace$ has been divided into $\subSearchSpaceCount$ non-overlapping regions $\searchSpace_1, \ldots, \searchSpace_{\subSearchSpaceCount}$, such that each region $\searchSpace_{\agentIndex}$
contains at least one base station, and assuming that the base stations in all regions can communicate with a central server $\centralServer$, then iteratively deploy agents $\subSearchSpaceCount$-at-a-time along paths $\agentPath_{1}^{(\roundIndex)} , \ldots, \agentPath_{\subSearchSpaceCount}^{(\roundIndex)} $ for rounds $\roundIndex = 1, 2, \ldots$ such that $\agentPath_{\agentIndex}^{(\roundIndex)}  \subseteq \searchSpace_{\agentIndex}^{(\roundIndex)} $ for each $\roundIndex$ (and all $\agentIndex$), and observe whether or not the agents survive to gather information about hazards to refine a hazard belief map $\environmentRandomVariable$. 
\end{problem}

When we consider Problem 3 in our experiments (Section~\ref{sec:experiments_and_results}), we assume  that exploration always begins and ends at each agent's respective base station $\baseStation_{\agentIndex}$ and there is only one base station per partition (${\baseStationCount = \subSearchSpaceCount}$). However, these requirements %that each agent start and end at the same base station 
can be relaxed, in general.

%---------------------------------------
%---------------------------------------
%---------------------------------------

\section{Path-Based Sensors with Bayesian Networks}
% \section{THE PROBABILISTIC GRAPHICAL MODEL APPROACH}
\label{sec:approach}
%---------------------------------------
%---------------------------------------
%---------------------------------------

In this section we present a methodology based on Bayesian networks for estimating hazard locations in the search space. We %start by revisiting the 
overview the hazard belief map update mechanism with path-based sensors~%employed in 
\cite{otte2021path}. % when the agent gets destroyed. 
Next, we propose a Bayesian network modeling method for sequential deployment of multiple robots to improve the information gathering rate. Subsequently, we %delve into the Bayesian network framework we propose and
discuss the integration of the Bayesian network method within the parallel multi-agent information gathering problem to improve the computational scalability. % introduced in~\cite{Srivastava.etal.DARS22}.

\subsection{Weighted-Average Path-Based Sensors}
\label{subsec:approach_old}
Upon observing a path-based sensor trigger ($\pathBasedSensorState_\agentIndex = 1$), the method %proposed 
in~\cite{otte2021path} computes the posterior belief map using a weighted-average technique. The method calculates the posterior belief maps produced by the mutually exclusive occurrence of the agent being destroyed at each cell along the path~$\agentPath_\agentIndex$. The final belief map (as calculated in~\cite{otte2021path}) is then found by combining the computed belief maps weighted by their relative likelihood of occurrence.
% \cite{otte2021path} describes a path-based sensor triggering as A path-based sensor triggering $(\pathBasedSensorState = 1)$, is described by the mutually exclusive events of path-based sensor triggering along each discrete cell $C$ of path $\agentPath$~\cite{otte2021path}. This is done by summing the belief maps resulting from each mutually exclusive event weighted by the relative likelihood.  
%
Thus, the belief map $\beliefMap{{\environmentRandomVariable}^{(\iterationIndex)}}$ at search round $\agentIndex$ is given by,
\begin{equation}
\label{eq:TASE_equation}
    \beliefMap{{\environmentRandomVariable}^{(\iterationIndex)}} = \sum_{\pathIndex = 1}^{\pathLength_\agentIndex} \Probability(\environmentRandomVariable^{(\iterationIndex)} | \environmentRandomVariable^{(\iterationIndex-1)}, \distruction_{\agentPath_\iterationIndex,\pathIndex} = 1)\Probability(\distruction_{\agentPath_\iterationIndex,\pathIndex} = 1),
\end{equation}
where $\distruction_{\agentPath_\iterationIndex,\pathIndex} = 1$ denotes the event of path-based sensor triggering at the $\pathIndex$-th cell of path $\agentPath_\iterationIndex$. %This approach~\eqref{eq:TASE_equation} does not conjecture which exact cell may have caused the path-based sensor triggering. Instead, it calculates a weighted average of mutually exclusive probability of each cell causing the trigger along a given path. 

\subsection{Bayesian Network Path-Based Sensors}
\label{subsec:pbs-bn}
%In this paper, w
We now propose a different methodology that uses a Bayesian network to incorporate the multi-universe %idea of~\cite{otte2021path} 
structure into a single joint-distribution. Our  methodology is comprised of three steps: (i)~observation; (ii)~inference; and (iii)~estimation. Based on the prior belief map of the environment, our method \emph{estimates} the new belief map after \emph{inferring} the origin of the \emph{observation} (i.e., location of agent destruction).

\begin{figure}[!t]

\centering

\noindent\begin{minipage}{\textwidth}
\begin{xy}
  \xyimport(100,100){\includegraphics[width=.77\textwidth]{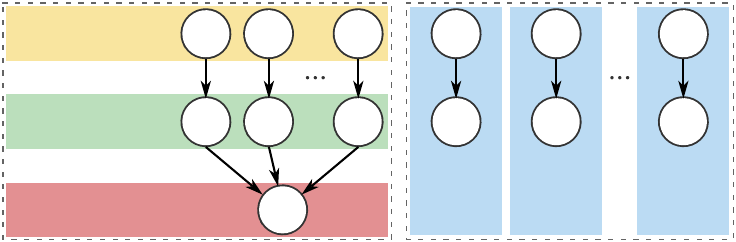}}
     ,(50,122)*{\text{Bayesian Networks for Path-Based Sensors}}
     ,(26.5,112)*{\text{\scriptsize Network when sensor is not triggered}}
     ,(26.5,104)*{\text{\scriptsize $\pathBasedSensorState = 1$ (robot is destroyed)}}
     ,(13.5,90)*{\text{\scriptsize Hazard belief}}   
     ,(13.5,83)*{\text{\scriptsize along path}}   
     ,(28,85.5)*{\text{$\environmentRandomVariable_1$}}  
     ,(36.5,85.5)*{\text{$\environmentRandomVariable_2$}}  
     ,(48.75,85.5)*{\text{$\environmentRandomVariable_\pathLength$}}  
     ,(13.5,54)*{\text{\scriptsize Destruction at}}   
     ,(13.5,47)*{\text{\scriptsize each cell}}  
     ,(28,49.5)*{\text{$\distruction_1$}}  
     ,(36.5,49.5)*{\text{$\distruction_2$}}  
     ,(48.75,49.5)*{\text{$\distruction_\pathLength$}}  
     ,(13.5,18)*{\text{\scriptsize Path-based sensor}}   
     ,(13.5,10)*{\text{\scriptsize observation}}  
     ,(38.5,12.75)*{\text{$\pathBasedSensorState$}}
     ,(26.5,-6)*{\text{\small (a)}}
     ,(77.5,112)*{\text{\scriptsize Network when sensor is triggered}}
     ,(77.5,104)*{\text{\scriptsize $\pathBasedSensorState = 0$ (robot survives)}}
     ,(62,85.5)*{\text{$\environmentRandomVariable_1$}}  
     ,(75.5,85.5)*{\text{$\environmentRandomVariable_2$}}  
     ,(93,85.5)*{\text{$\environmentRandomVariable_\pathLength$}}  
     ,(62,49.5)*{\text{$\distruction_1$}}  
     ,(75.5,49.5)*{\text{$\distruction_2$}}  
     ,(93,49.5)*{\text{$\distruction_\pathLength$}}  
     ,(62,32)*{\text{\scriptsize Bayesian}}   
     ,(62,23)*{\text{\scriptsize network}}   
     ,(62,16)*{\text{\scriptsize for}}   
     ,(62,8)*{\text{\scriptsize cell $\cellMapCell_1$}}   
     ,(75.5,32)*{\text{\scriptsize Bayesian}}   
     ,(75.5,23)*{\text{\scriptsize network}}   
     ,(75.5,16)*{\text{\scriptsize for}}   
     ,(75.5,8)*{\text{\scriptsize cell $\cellMapCell_2$}}
     ,(93.0,32)*{\text{\scriptsize Bayesian}}   
     ,(93.0,23)*{\text{\scriptsize network}}   
     ,(93.0,16)*{\text{\scriptsize for}}   
     ,(93.0,8)*{\text{\scriptsize cell $\cellMapCell_\pathLength$}}  
     ,(77.5,-6)*{\text{\small (b)}}
\end{xy}
\end{minipage}

    \caption{Bayesian networks used for inference based on path-based sensor triggering. (a) Bayesian network used when the path-based sensor is triggered (the robot is destroyed, ${\pathBasedSensorState = 1}$) along a path of length $\pathLength$. The \textit{observation} layer (red) indicates whether the path-based sensor was triggered. The {\textit{inference}} layer (green) speculates the most likely triggering cell based on the prior belief map. The \textit{estimation} layer (yellow) represents the belief of hazards $\beliefMap{\environmentRandomVariable}$. The causal flow of this Bayesian network starts with the prior belief in the {estimation} layer, causing agent's destruction in the {{inference}} layer, triggering the path-based sensor observation in the \textit{observation} layer. (b) The $\pathLength$ Bayesian networks used when the path-based sensor is not triggered (when the robot survives, $\pathBasedSensorState = 0$, it survives each of the $\pathLength$ cells along the path).}
    \label{fig:BN_Layers}
\end{figure}

Consider that an agent traverses a path $\agentPath_{\iterationIndex}=\langle \cellMapCell_1, \cellMapCell_2, \ldots \cellMapCell_{\pathLength}\rangle$ and never reaches its designated goal cell $\cellMapCell_{\pathLength}$. The problem can be modeled in a Bayesian network (Fig.~\ref{fig:BN_Layers}-(a)). The upper layer containing $\environmentRandomVariableSet_{\agentPath} = \environmentRandomVariable_1, \environmentRandomVariable_2, \ldots, \environmentRandomVariable_{\pathLength}$ is referred to as the \emph{estimation} layer, where $\environmentRandomVariableSet_{\agentPath}$ denotes the random variables of hazard existence at cells $\cellMapCell_1, \cellMapCell_2, \ldots, \cellMapCell_{\pathLength}$ along the path $\agentPath_{\iterationIndex}$. The  variables in the estimation layer are each connected to $\distructionSet_{\agentPath} = \distruction_1, \distruction_2, \ldots, \distruction_{\pathLength}$, where $\distructionSet_{\agentPath}$ denotes the random variables of destruction in cells $\cellMapCell_1, \cellMapCell_2, \ldots, \cellMapCell_{\pathLength}$. This layer is referred to as the \emph{inference} layer. The final layer is called the \emph{observation} layer. Every element in the inference layer is connected to a single variable $\pathBasedSensorState$ (i.e., observation) which represents the random variable for triggering a path-based sensor. % which is our . 

This \emph{simply-connected directed acyclic graph} (Fig~\ref{fig:BN_Layers}-(a)) is a Bayesian network with joint distribution equal to the sum of the product for all possible distributions. For brevity we omit subscript $i$ for the remainder of Section~\ref{subsec:pbs-bn}. % is given by 
\begin{equation}
\label{eq:BN_joint_distribution}
\Probability(\environmentRandomVariable, \distruction, \pathBasedSensorState)  
= \sum_{\substack{\beliefRealizationSet \in \{0, 1\}^{\mapHeight \times \mapWidth}\\ 
        \distructionRealizationSet \in \{0, 1\}^{\pathLength}\\ 
        \pathBasedSensorRealization \in  \{0, 1\}}}\ 
        \prod_{\substack{\otherPathIndex \in [1, \mapHeight \times \mapWidth] \\ 
        \pathIndex \in [1, \pathLength]}}
        \Probability(\environmentRandomVariable_\otherPathIndex = \beliefRealization_\otherPathIndex, \distruction_{\pathIndex} = \distructionRealization_{\pathIndex}, \pathBasedSensorState = \pathBasedSensorRealization)
\end{equation}
Given our assumption that hazard effects are local to each cell,~\eqref{eq:BN_joint_distribution} can be simplified by only considering the sequence of $\pathLength$ cells through which the path traverses, and not all ${\mapHeight \times \mapWidth}$ cells in the entire search space. 
The cell joint distribution is simplified 
%based on the domain knowledge of the Bayesian network 
to,
\begin{equation*}
\Probability(\environmentRandomVariable_{\pathIndex}, \distruction_{\pathIndex}, \pathBasedSensorState) = \Probability(\pathBasedSensorState = \pathBasedSensorRealization | \distructionSet_{\agentPath} =\distructionRealizationSet)\Probability(\distruction_{\pathIndex} = \distructionRealization_{\pathIndex} | \environmentRandomVariable_{\pathIndex} = \beliefRealization_{\pathIndex})\Probability(\environmentRandomVariable_{\pathIndex} = \beliefRealization_\pathIndex),
\end{equation*}
where $\Probability(\pathBasedSensorState = \pathBasedSensorRealization | \distructionSet_{\agentPath}=\distructionRealizationSet)$ is the likelihood of a path-based sensor trigger $\pathBasedSensorState = \pathBasedSensorRealization$  along the path $\agentPath$, $\Probability(\distruction_{\pathIndex} = \distructionRealization | \environmentRandomVariable_\pathIndex = \beliefRealization)$ is the likelihood of destruction $\distruction_{\pathIndex}$ given the existence of hazard $\environmentRandomVariable_\pathIndex = \beliefRealization_\pathIndex$, and $\Probability(\environmentRandomVariable_{\pathIndex} = \beliefRealization_{\pathIndex})$ is the prior belief of the hazard $\environmentRandomVariable_{\pathIndex}$ at cell $C_{\pathIndex}$ of path $\agentPath$. The likelihood $\Probability(\distruction_{\pathIndex} | \environmentRandomVariable_{\pathIndex})$ and the prior $\Probability(\environmentRandomVariable_\pathIndex)$ have previously been used in various Bayesian Filters for updating belief maps and are derived from the application of Bayes’ rule in similar environments. The likelihood term $\Probability(\pathBasedSensorState | \boldsymbol{\distruction}_{\agentPath})$ incorporates the \emph{inference layer} of the Bayesian network. % However, the likelihood $\Probability(\pathBasedSensorState | \boldsymbol{\distruction}_{\agentPath})$ incorporates the \emph{inference layer} of the Bayesian network, thereby weighting the joint distribution with the likelihood of triggering by each cell. 
This term estimates the likelihood of tripping the path-based sensor given a specific permutation of $\pathLength$ plausible causes. The computation of this likelihood function is different for each specific application of path-based sensors. 

In our case---where we assume the destruction of the agent as a path-based sensor triggering---the likelihood function results in,
\begin{equation}\label{eq:inference}
\begin{aligned}
\Probability(\pathBasedSensorState = 1|\distructionSet = \distructionRealizationSet) = \begin{cases}
\prod_{\pathIndex = 1}^{\otherPathIndex} \Probability(\distruction_{\pathIndex} = \distructionRealization_{\pathIndex}), &\textrm{if} \ \sum_{\pathIndex = 1}^{\pathLength} \distructionRealization_{\pathIndex} \leq 1\\
0, &\textrm{if}\  \sum_{\pathIndex = 1}^{\pathLength} \distructionRealization_{\pathIndex} > 1,
\end{cases}
\end{aligned}
\end{equation} 
where $\otherPathIndex$ is the path index of the cell $\cellMapCell_{\otherPathIndex}$ at which $\distruction_{\otherPathIndex} = 1$. Therefore, the likelihood function specifies the probability that the agent will be destroyed at $\cellMapCell_{\otherPathIndex}$ of $\agentPath_{\iterationIndex}$ while surviving the $\otherPathIndex - 1$ cells preceding it.

\begin{remark}
The conditions in \eqref{eq:inference} do not include repeated cell visiting along the path $\agentPath_{\iterationIndex}$. In other words, the path sequence $\agentPath_{\iterationIndex}~=~\langle \cellMapCell_1, \ldots, \cellMapCell_{\pathLength} \rangle$ is the same as the ordered set $\{\cellMapCell_1, \ldots, \cellMapCell_{\pathLength}\}$.  However, we allow the agent to have the flexibility to revisit cells during a path traversal. This alters the likelihood of destruction in a particular cell, as the agent may get destroyed at any possible instance of being located in that cell. Therefore, the new likelihood $\Probability(\pathBasedSensorState |\distructionSet )$ is calculated by summing the probability of each instance. Note that the Bayesian network shown in Fig.~\ref{fig:BN_Layers}-(a) assumes that each variable is independent and corresponds to a specific cell of the path. %Therefore, the changes to equation (3) are also made accordingly. 
\end{remark}

Given an observation of path-based sensor $\pathBasedSensorState = \pathBasedSensorRealization$, the posterior of $\cellMapCell_{\pathIndex}$ cell of path $\agentPath_{\iterationIndex}$ follows Bayes' rule,
\begin{equation}
\label{eq:posterior}
\Probability(\environmentRandomVariable_{\pathIndex} = 1 | \pathBasedSensorState = \pathBasedSensorRealization) \propto \sum_{\beliefRealizationSet,\distructionRealizationSet \in \{0, 1\}^{\pathLength}}\  \!\!\!\!\!\!  \Probability(\environmentRandomVariable_{\pathIndex} = 1, \distruction_{\pathIndex} = \distructionRealization, \pathBasedSensorState)\prod_{\otherPathIndex \ne \pathIndex}\Probability(\environmentRandomVariable_{\otherPathIndex} = \beliefRealization_{\otherPathIndex}, \distruction_{\otherPathIndex} = \distructionRealization, \pathBasedSensorState).
\end{equation}
This unnormalized probability is normalized in the standard Bayesian manner. The calculation of the posterior probability of cell $\environmentRandomVariable_{\pathIndex}$ given the path-based sensor observation $\pathBasedSensorState$ using \eqref{eq:posterior} is computationally expensive and becomes intractable as the path-length $\pathLength$ increases. The exact inference technique of %\textbf
{variable elimination}~\cite[Chapter~9]{koller2009probabilistic} is used to address this problem. This technique eliminates all the random variable of hazards $\environmentRandomVariable_{\otherPathIndex}$, where $\otherPathIndex \ne \pathIndex$, reducing the posterior probability of $\environmentRandomVariable_{\pathIndex}$ \eqref{eq:posterior} to,
\begin{equation}\label{eq:posterior_new}  
\Probability(\environmentRandomVariable_{\pathIndex} = 1 | \pathBasedSensorState = \pathBasedSensorRealization) \propto \nonumber
\sum_{\distructionRealizationSet \in \{0, 1\}}\ \Probability(\distruction_{\pathIndex} = \distructionRealization|\environmentRandomVariable_{\pathIndex} = 1)\Probability(\environmentRandomVariable_{\pathIndex} = 1)\Probability(\pathBasedSensorState|\distructionSet = \distructionRealizationSet) \prod_{\otherPathIndex \ne \pathIndex}\Probability(\distruction_j = \distructionRealization).
\end{equation}
Note that~\eqref{eq:posterior} and \eqref{eq:posterior_new} are relevant only when a path-based sensor is triggered, i.e. $\pathBasedSensorState = 1$. When $\pathBasedSensorState = 0$ the inference layer becomes redundant because each variable in the inference layer has a value of zero with probability equals 1, i.e. $ \distruction_{\pathIndex} = 0 \ \forall \pathIndex \in \{1,\ldots,\pathLength\}$. This certainty that ${\distruction = 0}$ allows the Bayesian network to be broken into $\pathLength$ separate Bayesian networks, as shown in Fig.~\ref{fig:BN_Layers}-(b). Thus, the joint distribution of $\environmentRandomVariable_{\pathIndex}$ and $\distruction_{\pathIndex}$ is simplified as %follows on the Bayesian network, % which yields,
\begin{equation}
    \Probability(\environmentRandomVariable_{\pathIndex} = 1, \distruction_{\pathIndex} = 0) = \Probability(\distruction_{\pathIndex} = 0|\environmentRandomVariable_{\pathIndex} = 1)\Probability(\environmentRandomVariable_{\pathIndex} = 1).
\end{equation}
In addition, the posterior of $\pathIndex$-th cell of path $\agentPath_{\iterationIndex}$ in the event of no path-based sensor is triggered takes the form of
\begin{equation}
\label{eq:posterior_survive}
    \Probability(\environmentRandomVariable_{\pathIndex} = 1 | \distruction_{\pathIndex} = 0) = \frac{\Probability(\distruction_{\pathIndex} = 0|\environmentRandomVariable_{\pathIndex} = 1)\Probability(\environmentRandomVariable_{\pathIndex} = 1)}{\Probability(\distruction_{\pathIndex} = 0)}.
\end{equation}

\subsection{Multi-Robot Bayesian Network Path-Based Sensors}
\label{subsec:approach_multi}
The Bayesian network path-based sensor formulation (presented in Section~\ref{subsec:pbs-bn})  can also be combined with the Distributed Entropy-weighted Voronoi Partition and Planner (DEVPP) framework described in \cite{Srivastava.etal.DARS22}. 
A distributed strategy utilizes multiple agents in parallel in non-overlapping regions of the search space. This is achieved in two steps: (i) entropy-based map partitioning; and (ii) local information-theoretic planning. 

In (i), the search space~$\searchSpace$ is partitioned into $\subSearchSpaceCount$ regions, one for each robot to be deployed during the next search round $\roundIndex$. The partitioning is accomplished by employing a weighted Voronoi partition mechanism with $\subSearchSpaceCount$ base stations $\baseStation_1, \ldots, \baseStation_{\subSearchSpaceCount}$,  serving as generator points. Let $\location_{\baseStation_{\agentIndex}}$ denote the position of $\baseStation_{\agentIndex}$
and let $\location_{\cellMapCell}$ denote the position of the center of cell $\cellMapCell$.
The entropy-weighted distance function used in \cite{Srivastava.etal.DARS22} to help construct the entropy-weighted Voronoi partitions yields,
$$
\distanceFunction(\location_{\cellMapCell}, \location_{\baseStation_{\agentIndex}}) = 
\relativeWeight(\hazardPresenceInCell,\environmentRandomVariable_{\searchSpace_{\agentIndex}^{(\roundIndex)}})
\,\,\,
\lvert\lvert(\location_{\cellMapCell} - \location_{\baseStation_{\agentIndex}})\lvert\lvert_1,
$$
where $\lvert\lvert \cdot \lvert\lvert_1$ is the $\mathrm{L1}$ norm and $\relativeWeight(\hazardPresenceInCell,\environmentRandomVariable_{\searchSpace_{\agentIndex}^{(\roundIndex)} })$ is the average entropy of the expected partition if cell $\cellMapCell \subset \searchSpace$ is added to subspace, and $\searchSpace_{\agentIndex^{(\roundIndex)}}$ is the weighted Voronoi region associated with base station $\baseStation_\agentIndex$,
\begin{equation*}
    \relativeWeight(\hazardPresenceInCell,\environmentRandomVariable_{\searchSpace_{\agentIndex}^{(\roundIndex)} }) = \frac{\Entropy(\hazardPresenceInCell)+\Entropy(\environmentRandomVariable_{\searchSpace_{\agentIndex}^{(\roundIndex)} })}{\mathrm{card}(\environmentRandomVariable_{\searchSpace_{\agentIndex}^{(\roundIndex)} })+1}.
\end{equation*}
Thus, during round $\roundIndex$ the subspaces $\searchSpace_{\agentIndex}^{(\roundIndex)}$ for each agent $\agentIndex$ are constructed to contain approximately equal Shannon information entropy according to the rule,
\begin{equation}
\label{eq:entropy_partition}
\searchSpace_{\agentIndex}^{(\roundIndex)} = \bigcup_{\cellMapCell \subset \searchSpace} ( \cellMapCell \ |\ \distanceFunction(\location_{\cellMapCell}, \location_{\baseStation_{\agentIndex}}) \leq  \distanceFunction(\location_{\cellMapCell}, \location_{\baseStation_{\agentOtherIndex}}), \forall\ \agentIndex \neq \agentOtherIndex ).
\end{equation}

In step (ii), each agent $i$ 
is assigned the task of finding an optimal path $\agentPath^{(\roundIndex)*}_{\agentIndex}$ in its respective local search space~$\searchSpace_{\agentIndex}^{(\roundIndex)}$. The objective is to maximize the expected information gain about hazards $\environmentRandomVariable$ given the observed path-based sensor measurements $\pathBasedSensorState_{\agentPath_{\agentIndex}}$ along path $\agentPath_{\agentIndex}$ at search round~$\roundIndex$. In contrast to the approach in \cite{Srivastava.etal.DARS22}, in the current paper each of these individual path planning problems is solved using Bayesian network modeling that is described in Section~\ref{subsec:pbs-bn}.

The two-step methodology is presented in Fig.~\ref{fig:concept_image}, where (i) a central entity decomposes the environment into search spaces using an entropy-weighted Voronoi partitioning~%assigned to each agent %in a centralized fashion~
(Fig.~\ref{fig:concept_image}-(a)); and (ii)~each agent computes an informative path in its assigned region in a distributed manner to maximize information about %environmental 
hazards in their assigned search space using the proposed Bayesian network-based planner (Fig.~\ref{fig:concept_image}-(b)). After each %At the end of a 
%search 
round, the centralized server updates the belief of the hazard state and assigns new search spaces with approximately equal information to each agent %to distribute the search effort among robots 
%using an entropy-weighted Voronoi partitioning~
(Fig.~\ref{fig:concept_image}-(c)). Then, the agents compute an informative path in their new search spaces to gather information about %environmental 
hazards~Fig.~\ref{fig:concept_image}-(d). Note that we also incorporate false-positive and false-negative sensor observations, making the model resilient to noisy data. %Over multiple deployments, the location of the hazards is refined.

Following the computation of local informative paths, each agent traverses the planned paths. Then, the base stations observe the survival of each agent (i.e., path-based sensor) and communicate the observations to a central server that updates the belief map of the environment. This process is repeated at the end of each search round, where the methodology is applied recursively to compute new local search spaces based on the posterior belief map and make new path-based sensor observations as illustrated in Fig.~\ref{fig:concept_image}.

% Bookmark for Alkesh
\section{Algorithms}
\label{sec:algorithms}
In this section, we present algorithms that use our Bayesian network modeling approach to address Problems \ref{ps:problem} and \ref{ps:multi} by estimating the posterior hazard belief map of an unknown environment recursively. We call our algorithm that addresses Problem \ref{ps:problem} ``Bayesian Network-based Information Theoretic Planner'' (\textsc{BNITP}), and our algorithm that addresses Problem \ref{ps:multi} ``Bayesian Network-based Distributed Entropy-Weighted Voronoi Partition and Planner'' (\textsc{BN-DEVPP}). 
Because much of the logic of \textsc{BNITP} is also used in \textsc{BN-DEVPP} to solve Problem \ref{ps:multi}, it is convenient to break the presentation of \textsc{BNITP} into a high-level outer loop (Algorithm~\ref{alg:outerLoop}) that invokes a subroutine (Algorithm~\ref{alg:mutual_info_planning}) once per iteration. 
While our presentation uses a maximum number of iterations $\iterationIndex_{\mathrm max}$, it is possible to set this to $\infty$ and/or to use some other stopping criterion. The outer loop of BNITP simply calls the BNITP repeatedly.

\begin{algorithm}[b!]
{\small\caption{: One Agent per Iteration  (\textsc{BNITP-loop})}\label{alg:outerLoop}
\noindent\small\textbf{Inputs}: 
hazard belief map prior $\beliefMap{\environmentRandomVariable^{(0)}}$, 
search space $\searchSpace$,
base station set  $\baseStationSet$,
maximum number of iterations $\iterationIndex_{\mathrm max}$\\
\textbf{Output}: $\beliefMap{\environmentRandomVariable^{(\iterationIndex_{\mathrm max})}}$
}
\begin{algorithmic}[1]\small
    \For{$\iterationIndex = 1$ to $\iterationIndex_{\mathrm max} $} 
        \State $\beliefMap{\environmentRandomVariable^{(\iterationIndex)}}$  $\gets$ \textsc{BNITP}-iteration ($\beliefMap{\environmentRandomVariable^{(\iterationIndex - 1)}}$, $\searchSpace$, $\baseStationSet$) ~{\color{gray}// Algorithm \ref{alg:mutual_info_planning}}
    \EndFor
\end{algorithmic}
\end{algorithm}

\begin{algorithm}[b!]
{\small\caption{: One Agent per Iteration  (\textsc{BNITP}-iteration)}\label{alg:mutual_info_planning}
\noindent\textbf{Inputs}: 
prior hazard belief map $\beliefMap{\environmentRandomVariable^{(\agentIndex-1)}}$, 
search space $\searchSpace$,
base station set  $\baseStationSet$\\
\textbf{Output}: posterior hazard belief map $\beliefMap{\environmentRandomVariable^{(\agentIndex)}}$
}
\begin{algorithmic}[1] \small
% \State $\agentPath_1 = {\color{blue}\texttt{calculateBNPath}}(Z^{(1)}, S)$
\State $\agentPath_{\agentIndex} \gets {\texttt{calculateBNPath}}(\environmentRandomVariable^{(\agentIndex-1)}, \searchSpace, \baseStationSet)$ ~{\color{gray}\,\,\,\,\,// Algorithm \ref{alg:calculate-BN-path}, contains difference vs.\ \cite{otte2021path}}
\State  observe $\pathBasedSensorRealization_\agentIndex$ after agent traverses path $\agentPath_{\agentIndex}$
\If{$\pathBasedSensorRealization_\agentIndex = 1$} 
    \State $\environmentRandomVariable^{(\agentIndex)} = {\texttt{BN\_PBSTrig}}(\environmentRandomVariable^{(\agentIndex-1)},\agentPath_{\agentIndex}, \pathBasedSensorRealization_\agentIndex)$~{\color{gray}\,\,\,\,\,\,  //  \eqref{eq:posterior}, key difference vs.\ \cite{otte2021path}}
\Else
    \State $\environmentRandomVariable^{(\agentIndex)} = {\texttt{BN\_NoTrig}}(\environmentRandomVariable^{(\agentIndex-1)}, \agentPath_{\agentIndex}, \pathBasedSensorRealization_\agentIndex)$~{\color{gray}\,\,\,\,\,\,\,\,\,  //  \eqref{eq:posterior_survive}}
\EndIf 
% \State \textbf{return} $Z^{(r)}$
\end{algorithmic}
\end{algorithm}

The implementation of a single \textsc{BNITP} iteration
is presented in~Algorithm~\ref{alg:mutual_info_planning}. We compute the relaxed optimal path $\agentPath_{\agentIndex}$ based on the prior hazard belief map $\environmentRandomVariable^{(\agentIndex-1)}$ using the subroutine $\texttt{calculateBNPath}$ (line~2). $\texttt{calculateBNPath}$ calculates the path as originally described in \cite{otte2021path}, except that path-based sensor updates are performed according to our new Baysian networked-based method method. We present $\texttt{calculateBNPath}$ in the appendix as Algorithm~\ref{alg:calculate-BN-path}. Next, we observe the realization of the path-based sensor, where $\pathBasedSensorRealization_\agentIndex=1$ if the agent is destroyed and $\pathBasedSensorRealization_\agentIndex=0$ if the agent survives from the traversed path (line 3). % represents the output $\pathBasedSensorState$ of the path-based sensor if the agent traverses the path obtained in line 2. 
When the path-based sensor is triggered ($\pathBasedSensorRealization_\agentIndex = 1$), the posterior hazard belief map $\environmentRandomVariable^{(\agentIndex)}$ is updated using~\eqref{eq:posterior}. %, as discussed in Section~\ref{sec:approach}. 
Similarly, if the path-based sensor is not triggered ($\pathBasedSensorRealization_\agentIndex = 0$), then the posterior hazard belief map $\environmentRandomVariable^{(\agentIndex)}$ is updated using~\eqref{eq:posterior_survive}. The use of %Equation 
\eqref{eq:posterior} on line 4 is a key difference versus \cite{otte2021path}.

%The subroutine \texttt{calculateBNPath} (Algorithm~\ref{alg:calculate-BN-path}) extends the function \texttt{calculatePath}, where  \texttt{calculatePath} was originally presented in  \cite[Algorithm 2]{otte2021path}. Differences between our version and the version presented in \cite[Algorithm 2]{otte2021path} are highlighted in {\color{blue}blue} font; in particular, the expected hazard belief $Z$ is now calculated based on equations \eqref{eq:posterior} and \eqref{eq:posterior_survive}, respectively. 

\begin{algorithm}[tbh]
\caption{: $\subSearchSpaceCount$-Agents per Iteration \textsc{BN-DEVPP}}\label{alg:entropy-weighted-voronoi-partitioning-and-planning}
%\hspace*{\algorithmicindent} 
{\noindent\small\textbf{Inputs}: 
hazard belief map prior $\environmentRandomVariable^{(0)}$, 
search space $\searchSpace$,
number of regions $\subSearchSpaceCount$,
base station set $\baseStationSet$,
maximum number of rounds $\roundIndex_{\textrm{max}}$\\
\textbf{Output}: $\environmentRandomVariable^{(\roundIndex_{\textrm{max}})}$
}
\begin{algorithmic}[1]\small
\For{$\roundIndex = 1$ to $\roundIndex_{\textrm{max}}$}
    \State calculate $\searchSpace_{1}^{(\roundIndex)}, \ldots, \searchSpace_{\subSearchSpaceCount}^{(\roundIndex)}$ {\color{gray} \,\,\,\,\,\,\,\,\,\,\,\,\,\,\,\,\,\,\,\,\,\,\,\,\,\,\,\,\,\,\,\,\,\,\,\,\,\,\,\,\,\,\,\,\,\,\,\,\,\,\,\,\,\,\,\,\,\,\,\,\,\,\,\,\,\,\,\,\,\,  // on server $\centralServer$, %Equation 
    \eqref{eq:entropy_partition}}
    \State $\environmentRandomVariable_{\searchSpace_{1}^{(\roundIndex)}}^{(\roundIndex - 1)} \cup \ldots \cup \environmentRandomVariable_{\searchSpace_{\subSearchSpaceCount}^{(\roundIndex)}}^{(\roundIndex - 1)} \gets \environmentRandomVariable^{(\roundIndex - 1)}$ {\color{gray} \,\,\,\,\,\,\,\,\,\,\,\,\,\,\,\,\,\,\,\,\,\,\,\,\,\,\,\,\,\,\,\,\,\,\,\,\,\,\,\,\,\,\,\,\,\,\,  // on server $\centralServer$}
    \State broadcast $\searchSpace_{\agentIndex}^{(\roundIndex)}$ and $\environmentRandomVariable_{\searchSpace_{\agentIndex}^{(\roundIndex)}}^{(\roundIndex - 1)}$ from $\centralServer$ to $\baseStation_{\agentIndex}$  {\color{gray}\,\,\,\,\,\,\,\,\,\,\,\,\,\,\,\,\,\,\,\,\,\,\,\,\,  // on server $\centralServer$}
    \For{$\agentIndex = 1,\ldots, \subSearchSpaceCount$}{\,\,\,\,\,\,\,\,\,\,\,\,\,\,\,\,\,\,\,\,\,\,\,\,\,\,\,\,\,\,\,\,\,\,\,\,\,\,\,\,\,\,\,\,\,\,\,\,\,\,\,\,\,\,\,\,\,\,\,\,\,\,\,\,\,\,\,\,\,\,\,\,\,\,\,\,\,\,\,\, \color{gray} // in parallel}
        \State $\environmentRandomVariable_{\searchSpace_{\agentIndex}^{(\roundIndex)}}^{(\roundIndex)} \gets$ {$\textsc{BNITP-iteration}$}$(\environmentRandomVariable_{\searchSpace_{\agentIndex}^{(\roundIndex)}}^{(\roundIndex - 1)},\baseStationSet,\searchSpace_i^{(\roundIndex)})$ {\color{gray} \,\, // difference vs. \cite{Srivastava.etal.DARS22} }
        \State {communicate $\environmentRandomVariable_{\searchSpace_{\agentIndex}^{(\roundIndex)}}^{(\roundIndex)}$ from $\baseStation_{\agentIndex}$ to $\centralServer$}
    \EndFor
    \State $\environmentRandomVariable^{(\roundIndex)} \gets \environmentRandomVariable_{\searchSpace_{1}^{(\roundIndex)}}^{(\roundIndex)} \cup \ldots \cup \environmentRandomVariable_{\searchSpace_{\subSearchSpaceCount}^{(\roundIndex)}}^{(\roundIndex)}$ {\color{gray} \,\,\,\,\,\,\,\,\,\,\,\,\,\,\,\,\,\,\,\,\,\,\,\,\,\,\,\,\,\,\,\,\,\,\,\,\,\,\,\,\,\,\,\,\,\,\,\,\,\,\,\,\,\,\,\,\,\,\,\,  // on server $\centralServer$}
\EndFor
\end{algorithmic}
\end{algorithm}

Algorithm~\ref{alg:entropy-weighted-voronoi-partitioning-and-planning} outlines the details of the BN-DEVPP method. After each search round $\roundIndex$, the centralized server $\centralServer$ initiates the algorithm by transmitting the prior belief map to all base stations $\baseStation_1, \ldots, \baseStation_{\subSearchSpaceCount}$. Subsequently, %based on the DEVPP methodology~\cite{Srivastava.etal.DARS22}, 
the central server computes the local search spaces $\searchSpace_{\agentIndex}$ for each agent $\agentIndex$, which are then communicated to their respective base stations. Each agent $\agentIndex$ employs the \textsc{BNITP} routine (Algorithm~\ref{alg:mutual_info_planning}) to independently compute an information path that maximizes information gain within their local search space. %This distributed approach allows each agent to make informed decisions based on their own local information, improving the overall exploration and hazard mapping capabilities of the BN-DEVPP algorithm.

\section{Experiments and Results}
\label{sec:experiments_and_results}
In this section, we report results from numerical experiments %that we ran 
to evaluate the proposed methodologies. We compare the proposed techniques with~\cite{otte2021path} for sequential and with~\cite{Srivastava.etal.DARS22} for parallel deployment of multi-robot systems. % the multi-robot case. % in a spatial environment that contains hazards~$Z$ in unknown locations. %We first describe the environment in which the experiments were conducted and then discuss their outcomes.

\textbf{\emph{Experimental Setup}}: We consider a discrete spatial environment of dimension $\mapHeight \times \mapWidth = 15 \times 15$ cells that contain $7$ hazards in unknown locations. Our task is to detect the location of the hazards by deploying multiple robots, either sequentially and/or in parallel. A cell in the environment $\cellMapCell$ is either empty ($\environmentRandomVariable_\cellMapCell = 0$) or contains a hazard ($\environmentRandomVariable_\cellMapCell = 1$) that threatens to destroy the agent, i.e., trigger the path-based sensor. The agent's movement in the environment is determined by a $9$-grid connectivity. This means that each agent can select the next step either by transitioning to any of its $8$-neighboring cells or by remaining in the same cell. %In the experiments conducted, w
We consider two environments with \emph{hazard lethality} of $70\%$ and $90\%$. {Hazard lethality} refers to the likelihood of an agent's destruction if it reaches a cell $C$ that is occupied by a hazard $\hazardPresenceInCell = 1$. %; i.e., the hazard lethality is the likelihood that governs the true positive of a path-based sensor. 
Each agent has a malfunction probability of $5\%$ %when taking a step %the 
(false-positive) %per step of a path-based sensor triggering. The likelihood of an agent surviving a cell $C$ with a hazard ($\hazardPresenceInCell = 1$) is considered to be $5\%$, the false-negative per step of a path-based sensor.
and $5\%$ chances of surviving a cell $C$ with a hazard %$\hazardPresenceInCell= 1$~
(false-negative).
We conduct Monte Carlo experiments in simulation to test the efficacy and robustness of the proposed methods, 15 trials are performed for each combination of method and number of agents deployed in parallel.

\begin{figure}[!t]
    \centering
    \includegraphics[width=9cm]{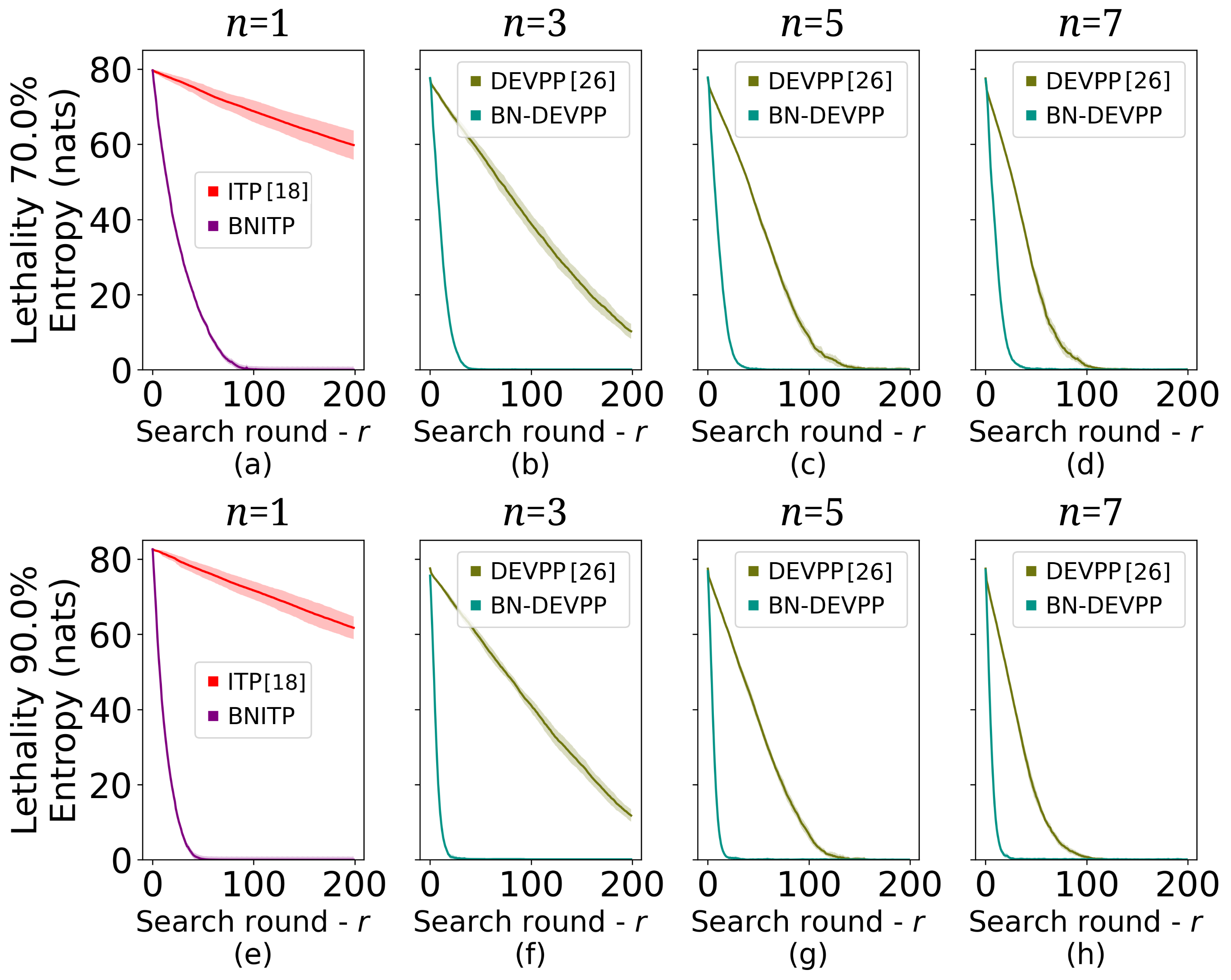}
    \caption{Comparison of information entropy across search rounds for different methods: (a) \textsc{BNITP} vs. \textsc{ITP}~\cite{otte2021path} for 70\% lethality; (b)-(d) \textsc{BN-DEVPP} vs. \textsc{DEVPP}~\cite{Srivastava.etal.DARS22} for 70\% lethality and $\subSearchSpaceCount=3,5,7$ respectively; (e) \textsc{BNITP} vs. \textsc{ITP}~\cite{otte2021path}; and (f)-(h) \textsc{BN-DEVPP} vs. \textsc{DEVPP}~\cite{Srivastava.etal.DARS22} for 90\% hazard lethality and $\subSearchSpaceCount=3,5,7$ respectively.}
%     (a) Comparison of the proposed BN-DEVPP algorithm and DEVPP~\cite{Srivastava.etal.DARS22} algorithm in terms of reduction in information entropy of the environment with $90\%$ adversary lethality when 3 agents are deployed simultaneously. 
% (b) Comparison of the proposed BN-DEVPP algorithm and DEVPP~\cite{Srivastava.etal.DARS22} algorithm in terms of reduction in information entropy of the environment with $90\%$ adversary lethality when 5 agents are deployed simultaneously. 

    \label{fig:entropy_comparison_MRS}
\end{figure}

\newcommand{\hpd}{@{\hspace{5pt}}}  % horizontal table padding

\begin{table}
\centering

\caption{Average Number of Agents Destroyed, for Various Methods and $\subSearchSpaceCount$}
\noindent\begin{minipage}{0.4\textwidth}
\begin{tabular}{| \hpd c \hpd | \hpd c \hpd | \hpd c \hpd |}
\multicolumn{3}{ c }{Lethality Rate $70\%$} \\
\hline
$\subSearchSpaceCount$ & DEVPP~\cite{Srivastava.etal.DARS22} & BN-DEVPP \\
\hline
    1 &  $>1000$~\cite{otte2021path} & \textbf{60.1} \\
    3 & 464.9 & \textbf{54.7} \\
    5 & 294.1 & \textbf{71.7}  \\
    7 & 261.2   & \textbf{86.6} \\
\hline
\end{tabular}
\label{tab:losses}
\end{minipage}
\hspace{0.5cm}
\begin{minipage}{0.4\textwidth}
\begin{tabular}{|\hpd c \hpd | \hpd c \hpd | \hpd c \hpd |}
\multicolumn{3}{ c }{Lethality Rate $90\%$} \\
\hline
$\subSearchSpaceCount$ & DEVPP~\cite{Srivastava.etal.DARS22} & BN-DEVPP \\
\hline
    1 &  $>1000$~\cite{otte2021path}  &  \textbf{25.2} \\
    3 & 478.3 & \textbf{26.6}  \\
    5 & 287.9 & \textbf{40.3} \\
    7 & 245.6  & \textbf{51.2}  \\
\hline
\end{tabular}
\label{tab:losses}
\end{minipage}
\end{table}

\textbf{\emph{Results and Discussion}}: We conduct three experiments to study the efficacy of the proposed methodologies. In the first experiment, we compare the proposed methodologies with~\cite{otte2021path} and~\cite{Srivastava.etal.DARS22}. Fig.~\ref{fig:entropy_comparison_MRS} illustrates the comparison of information entropy (i.e., negative uncertainty) of the environment in subsequent search rounds for all methods. %Note that, f
For the case of sequential agent deployment ($\subSearchSpaceCount=1$), we compare our \textsc{BNITP} algorithm with the ITP algorithm~\cite{otte2021path}, while for %the case of 
parallel deployment with multiple agents ($\subSearchSpaceCount>1$), we compare our \textsc{BN-DEVPP} with~\textsc{DEVPP}~\cite{Srivastava.etal.DARS22}. The results demonstrate that as the number of agents increases, the information entropy reduces at a faster rate in both environments with $70\%$ and $90\%$ lethality. %The steeper descent of the proposed methodologies indicates that our proposed methodologies gather information about hazard locations at a significantly higher rate.
This means that the proposed methodologies (\textsc{BNITP} and \textsc{BN-DEVPP}) 
gather significantly more information about hazard locations in the first $r=50$ search rounds. %at a significantly higher rate.
These findings suggest that the Bayesian network-based approach is more efficient at detecting hazards in communication-denied environments.

%inferring hazard belief maps, leading to better information gathering.

% In the second set of experiments, we deployed multiple agents simultaneously and compared the performance of \textsc{BN-DEVPP} with \textsc{DEVPP}\cite{Srivastava.etal.DARS22}. Fig.\ref{fig:entropy_comparison_MRS}-(a) and Fig.~\ref{fig:entropy_comparison_MRS}-(b) illustrate the reduction of information entropy in the environment when deploying 3 agents and 5 agents simultaneously, respectively. The results show that \textsc{BN-DEVPP} gathers more information about the location of hazards within the same number of search rounds. Additionally, it is worth noting that increasing the number of simultaneous agent deployments significantly reduces the number of search rounds required to reduce the information entropy to nearly zero.

In the second experiment, we observe the number of agents lost in the hazardous environment until the information entropy decreased to $10\%$ of its initial value at search round $\roundIndex=0$. Table~\ref{tab:losses} presents the average number of agents lost during the experiment. Our proposed methodology exhibits a notable reduction in the number of agents lost compared to previous methods. In the environment with $70\%$ hazard lethality, we observe a $79.12\%$ average reduction in the number of agents lost (averaged over $\subSearchSpaceCount=3,5,7$), % factor of $4.78$
while in the environment with $90\%$ lethality, we observe a reduction of $88.32\%$ on average in agent destruction.

\begin{figure}[!tb]
\begin{minipage}[t]{3.25cm}
    \caption{Computation time per agent for different $\subSearchSpaceCount$. By definition ${\subSearchSpaceCount=1}$ for BNITP and ${\subSearchSpaceCount>1}$ for BN-DEVPP. Results for 70\% (Left) and 90\% (Right) lethal hazards. Box-plots show statistics over search rounds.} \label{fig:runtime}
\end{minipage}
\hspace{.05cm}
\begin{minipage}[t]{8.7cm}
\centering

\vspace{.2cm}

Computation time per agent

\vspace{.2cm}

    \includegraphics[width=8.7cm]{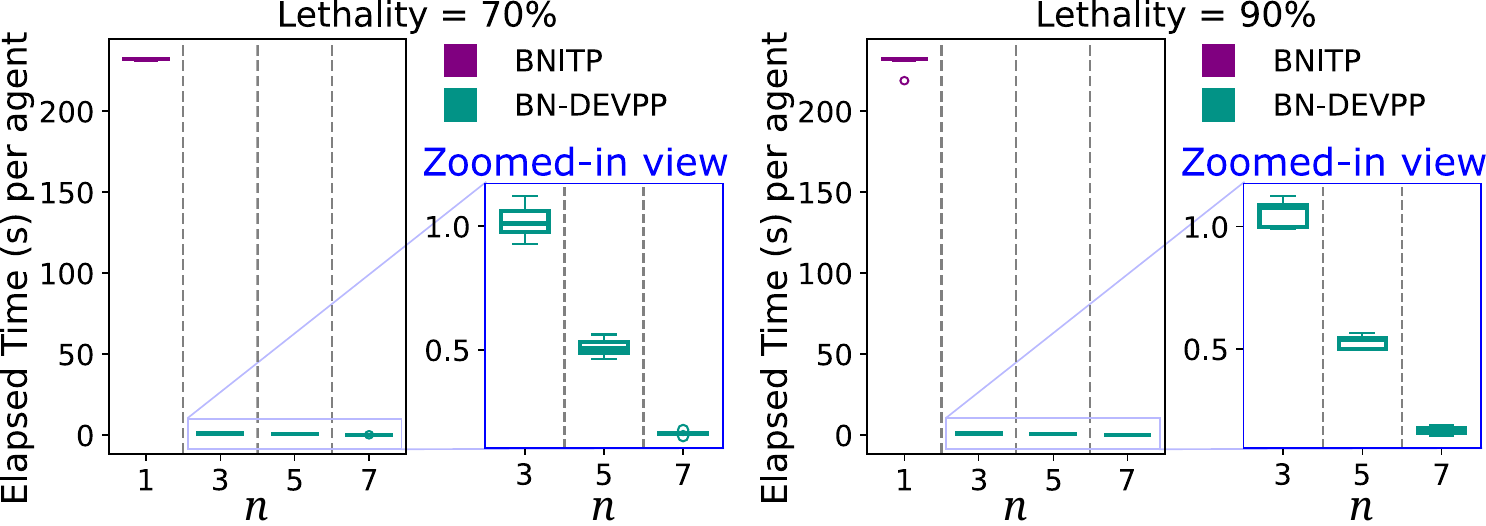}

\vspace{-.2cm}

{\scriptsize num agents deployed per round  \hspace{0.5cm} num agents deployed per round}
\end{minipage}

\end{figure}

The third experiment investigates the time required per agent to compute a path for a search round. The findings are illustrated in boxplots in Fig.~\ref{fig:runtime}. %It is noted that, 
We observe that increasing the number of agents leads to a reduction in the computation time for path planning. This scalability in computation time makes the algorithm suitable for real-time implementation. %These findings highlight the proposed method's ability 
The results reveal the ability of the proposed method to accelerate the information gathering process, minimize agent losses, and enable scalable multi-agent deployment. %, thereby demonstrating its effectiveness in real-time scenarios.

\section{Conclusion}
\label{sec:conclusion}
This paper presents a new Bayesian network modeling approach for path-based sensor updates. We introduce two algorithms, \textsc{BNITP} and \textsc{BN-DEVPP}, and through a series of experiments considering hazard detection in communication-denied environments, we compare the proposed method to prior work. We find that the proposed Bayesian network approach outperforms existing methods by achieving faster and more efficient information gathering while reducing the number of agents lost. In particular, the new method outperforms the sequential method from \cite{otte2021path} in the sequential case, and $n$-robot-at-a-time method from \cite{Srivastava.etal.DARS22} in the parallel case. 
We verify that a result from \cite{Srivastava.etal.DARS22}, that deployment of multiple agents in parallel improves scalability of path-based sensors algorithms, also holds for the new method.
By distributing the information gathering tasks among multiple agents, the system benefits from parallelization and leads to improved efficiency and scalability, as the agents can cover larger areas without task duplicity. 
%
% during the information gathering of lethal hazards. 
Overall, the Bayesian network formulation provides an efficient and robust framework for path-based sensors.

\section{Acknowledgments}

This work is supported by the Maryland Robotics Center and the Office of Naval Research (ONR) via grant N$0001420$WX$01827$~and~N$00014$-$20$-$1$-$2712$. The views, positions, and conclusions contained in this document are solely those of the authors and do not explicitly represent those of ONR.

\appendix

\section*{Appendix}

{\bf The indices $i,j,r$ used in the appendix match the notation used in~\cite{otte2021path} and they have different meanings than elsewhere in the current paper.}

Algorithm~\ref{alg:calculate-BN-path} presents pseudocode for \texttt{calculateBNPath} which is a modified version of Algorithm~2 appearing in \cite{otte2021path}. It performs a reverse search through a

\noindent\begin{minipage}[t]{0.4\textwidth}

graph $(\nodeSet_{\searchSpace\times\mathbb{T}}, \edgeSet_{\searchSpace\times\mathbb{T}})$ describing
the connectivity of $\searchSpace$ across time $\mathbb{T}$. A particular node ${\nu_i} \in \nodeSet_{\searchSpace\times\mathbb{T}}$ represents visiting cell $\cellMapCell$ at a particular time $\timeIndex$. 
The edge $({\nu_i},{\nu_j}) \in \edgeSet_{\searchSpace\times\mathbb{T}}$ connects node ${\nu_i}$ to ${\nu_j}$ (necessarily moving forward through time). The notation  $\agentPath_r \gets ({\nu_i},{\nu_j}) + \agentPath_{{\nu_j}}$ indicates that the subpath  $\agentPath_r$ is constructed by prepending edge $({\nu_i},{\nu_j})$ to the front of  subpath $\agentPath_{{\nu_j}}$, where $\agentPath_{{\nu_j}}$ starts at node ${\nu_j}$ and ends at the node associated with the goal cell $C_{goal}$. 
The variable $h$ is used to indicate the running expected information gain estimate, e.g., $h_{{\nu_i}}$ is the most expected information gained that has been found (so far) from any subpath starting from  ${\nu_i}$ and ending at the goal.

As described in \cite{otte2021path}, finding the true optimal path requires looking at all possible histories of reaching each node, which becomes intractable, even for small~$\searchSpace$. As in \cite{otte2021path}, tractability

\end{minipage}
\hspace{.05cm}
\begin{minipage}[t]{0.57\textwidth}

\vspace{-.8cm}

\begin{algorithm}[H] % Otte:  H fixes outer par loop error
\caption{\texttt{calculateBNPath}}\label{alg:calculate-BN-path}
{\noindent\small\textbf{Inputs}: prior hazard belief map $\beliefMap{\environmentRandomVariable^{(\iterationIndex-1)}}$, search space $\searchSpace$, nodes~$V_{\searchSpace \times \timeDimension}$\\
\textbf{Output}: path $\agentPath_r$ 
}
\begin{algorithmic}[1]\small
\For {\textbf{all} ${\nu_i} \in \nodeSet_{\searchSpace \times \timeDimension}$}
\State $h_{{\nu_i}} \rightarrow -\infty$
\EndFor
\State \texttt{InsertFIFOQueue}($C_{goal}$)

\While {${\nu_i} \gets \texttt{PopFIFOQueue}$}
    \For{\textbf{all} (${\nu_i},{\nu_j}) \in \edgeSet_{\mathrm{S\times \timeDimension}}$}
        \State $\agentPath_r \gets ({\nu_i},{\nu_j}) + \agentPath_{{\nu_j}}$
% \Comment Identical to \cite[Algortihm 2]{otte2021path} with line 10, 11 replaced by

% \Statex \small{Identical to \cite[Algorithm 2]{otte2021path} with line 10, 11 replaced by}
% \setcounter{ALG@line}{9}
% \Statex \ \ \ \vdots
\State {\color{gray}  // Equation \eqref{eq:posterior}, key diff. vs. \cite{otte2021path}: }

        \hspace{0.2cm} $(\beliefMap{\environmentRandomVariable_{\textrm{live}}}, p_{\agentPath_r}^{\textrm{alive}}) \gets$

        \hspace{1.1cm} ${\texttt{BN\_PBSTrigger}}(\beliefMap{\environmentRandomVariable^{(r-1)}},\agentPath_r,1)$

\State  {\color{gray} // Equation \eqref{eq:posterior_survive}:}

        \hspace{0.2cm} $\beliefMap{\environmentRandomVariable_{\textrm{killed}}} \gets$

        \hspace{1.7cm} ${\texttt{BN\_NoTrigger}}(Z^{(r-1)},\agentPath_r, 0)$

\State ${h_{this}} \gets  p_\agentPath^{alive}\Entropy({\environmentRandomVariable_{live}})$
        
        \hspace{1cm} $+ (1 -p_\agentPath^{alive})\Entropy({\environmentRandomVariable_{killed}})$
        \If{${h_{this}} > h_{{\nu_i}}$}
            \State $\agentPath_{{\nu_i}} \gets \agentPath_r$
            \State $h_{{\nu_i}} \gets h_{this}$
        \EndIf
        \For{\textbf{all} $({\nu_k}, {\nu_i}) \in \edgeSet_{\searchSpace\times\mathbb{T}}$}
            \If{ \textbf{not} \texttt{InQueue}(${\nu_k}$)}
                \State \texttt{InsertFIFOQueue}(${\nu_k}$)
            \EndIf
        \EndFor
    \EndFor
\EndWhile
\State $w_{start} \gets \mathrm{arg\ min}_{w \in W_{start}} h_w$
\State $\agentPath_r \gets \agentPath_{w_{start}}$
% \Statex \ \ \  \vdotssummary
% \setcounter{ALG@line}{20}
\State \textbf{return} $\agentPath_r$
\end{algorithmic}
\end{algorithm}

\end{minipage}

\noindent is achieved by a relaxation that iteratively considers only the set of subpaths that can be created by connecting nodes at time slice $\timeIndex$ to the best subpaths found starting at time slice $\timeIndex + 1$. We are also able to use a first-in-first-out queue because paths must move forward through time.

 \bibliographystyle{splncs04}
 \bibliography{mybib}

\end{document}